\documentclass[lettersize,journal]{IEEEtran}
\usepackage{amsmath,amsfonts}
\usepackage{algorithmic}
\usepackage{algorithm}
\usepackage{array}
\usepackage[caption=false,font=normalsize,labelfont=sf,textfont=sf]{subfig}
\usepackage{textcomp}
\usepackage{stfloats}
\usepackage{url}
\usepackage{verbatim}
\usepackage{graphicx}
\usepackage{hyperref}
\usepackage{cite}
\usepackage{multirow}
\usepackage{booktabs}
\usepackage{caption}
\usepackage[table]{xcolor}
\usepackage{colortbl}
\newcommand{\mycc}[1]{\cellcolor{gray!30}#1}

\begin{document}
%option 1
%\title{Multimodal Models for Cross-Domain Presentation Attack Detection on ID Documents}

\title{From Vision to Text: A Compact Multimodal Approach for Robust, Cross-Domain Presentation Attack Detection on ID Cards}

\author{
    Qingwen Zeng,
    Juan~E.~Tapia,~\IEEEmembership{Senior~Member,~IEEE,},
    Sneha Das,~\IEEEmembership{Member,~IEEE,}
    ~Christoph~Busch,~\IEEEmembership{Fellow~Member,~IEEE}%
\thanks{Juan E. Tapia and Christoph Busch are with the da/sec-Biometrics and Security Research Group, Hochschule Darmstadt, Germany, e-mail: \{\href{juan.tapia-farias@h-da.de}{juan.tapia-farias}, \href{christoph.busch@h-da.de}{christoph.busch}\}@h-da.de.}%
\thanks{Qingwen Zeng and Sneha Das are from Technical University of Denmark (DTU), Denmark, email: \href{s232892@student.dtu.dk}{s232892@student.dtu.dk, sned@dtu.dk}.}%

\thanks{Corresponding author: Juan E Tapia. Manuscript received Month DD, YYYY; revised Month DD, YYYY.}}

\markboth{Journal of \LaTeX\ Class Files,~Vol.~14, No.~8, August~2021}%
{Shell \MakeLowercase{\textit{et al.}}: A Sample Article Using IEEEtran.cls for IEEE Journals}


\maketitle

\begin{abstract}
Cross-domain shifts challenge Presentation Attack Detection (PAD) on ID Cards, given the restricted data available due to privacy concerns. This work proposes a compact multimodal model, based on new generative and discriminative blocks, which combines visual and textual data for PAD on genuine and synthetic ID images. While multimodal models exhibit strong generalisation after supervised fine-tuning, they fail in zero-shot settings. Our findings underscore that model capacity and real-world data are essential for reliable PAD, while existing synthetic datasets may not reflect real-world challenges. We argue for a re-evaluation of synthetic data as a benchmark and emphasise the need for more realistic, diverse datasets to advance PAD research.
\end{abstract}


\section{Introduction}
\IEEEPARstart{A} multimodal model is an Artificial Intelligence (AI) network that can incorporate, process, and generate information from multiple data types such as text, image, and audio. These inputs can be integrated and analysed to enhance human comprehension and interaction, rather than relying on a single format~\cite{Survey-FM}. Current multimodal models employ Vision Transformers (VT) networks and encoders/decoders to extract features or tokens from the data. As output, image and text explanations or descriptions are expected for images, texts, or audio inputs. In addition, unimodal models have been used to present a partial model that considers only image information for training the system \cite{tapia2025FM}. 

These models are supported by advanced algorithms that are often developed by large technology companies, such as Meta, OpenAI, and Tesla, among others \cite{Survey-FM, Survey-explainability}. The amount of data needed to train these algorithms is huge, enabling generalization capabilities across multiple domains. Furthermore, these models provide new insights into identifying and exploring downstream models dedicated to specific tasks that cannot generalize to other domains. Because of data protection restrictions, these datasets contain sensitive information, including demographic data, names, nationalities, biometric data, and health information.

One such case that requires specific downstream models is the Presentation Attack Detection (PAD) on ID Cards. The PAD on ID Cards occurs in a remote onboarding scenario, which raises a challenge for detecting various types of tentative presentation attacks and preventing impersonation and ID substitution for economic gain. In the onboarding enrollment process, the subject is required to provide a selfie and a photo of their ID Card to verify their identity in an unsupervised task.

Most PAD-on-ID-Card models are trained on a few ID Card countries due to the limited availability of datasets, and as a result, they cannot generalize to new ID Card countries or to cross-country or unknown scenarios. This challenge limits companies' ability to enter or expand into new markets under the PAD system. 

Motivate for this limitation and the huge data used for training state-of-the-art models such as ChatGPT, Qwen, Deepseek, SmolVLM2 and many others \cite{Survey-FM}. We believe that these models can be adapted to the PAD ID Card task to achieve generalization capabilities. Our focus is on developing an {\it efficient} model that achieves good performance while balancing model size, parameter count, and low error rate. Based on that, we evaluated a deep learning model, an unimodal model, and multimodal models.

The main contributions of this work are:
\begin{itemize}
    \item Deep learning, Unimodal, and Multimodal models were explored to assess the generalization capabilities in multiple countries.
    \item A compact SmolVLM2-based multimodal PAD framework was adapted, including a \textit{Generative} and \textit{Discriminative} structure to improve PAD on ID Cards.
    \item A extensive cross-dataset evaluation was performed considering real ID Card datasets and three different synthetic passports; in total, five countries were evaluated.
    \item The code will be available for reproducibility in GitHub (upon acceptance).
\end{itemize}

The remainder of the article is organized as follows: Section~\ref{sec:RLW} summarizes related works on PAD. Section \ref{sec:dataset} describes the dataset used.  Section~\ref{sec:Method} describes the proposed method. Section~\ref{sec:metrics} describes the metrics. Section~\ref{sec:EAR} describes the experiments and results obtained from the proposed method; finally, Section~\ref{sec:Ana}, and~\ref{sec:conclusion} discusses the results and conclusion.

\section{Related Work}
\label{sec:RLW}

Today, PAD on ID Cards has been increased in response to data scarcity and various associated challenges in this area~\cite{Survey-PAD-IDCards,gonzalez_forged_2025}, such as a lack of bona fide and no public datasets. Several authors have developed algorithms for PAD based on Deep learning models~\cite{gonzalez_hybrid_2021, gonzalez_forged_2025} and have also used generative adversarial networks (GANs) to generate synthetic attacks and simulated bona fides. 

Markham et al. ~\cite{markham_open-set_2024} generated simulated print and screen attack ID cards using CycleGAN and Pix2Pix~\cite{pix2pix2017}. The authors utilizes open-source datasets of video clips that show presentations of ID documents from fake subjects. The goal is to determine whether incorporating synthetic presentation attack samples into the training set, rather than real samples, yields comparable PAD performance.

Benalcazar et al.~\cite{benalcazar_synthetic_2023} generated synthetic attacks based on GAN and template-generated images from the bona fide distribution noise as data supplements for bona fide images. The experiments showed a positive impact of this approach; it would be a cost-effective way to obtain new data without requiring new subjects to provide sensitive data. However, the quality of the final images should be improved.

Further, Guam et al.~\cite{guan_idnet_2024} generated a synthetic ID Card called ID-Net. ID-Net contains~800K images generated using Stable Diffusion models, including driver's licenses and passports. They combined quality-driven hyper-parameter tuning using Bayesian optimization and recent breakthroughs in generative AI for generating representative document templates and identities.

A similar synthetic ID Card dataset, called FantasyID, was developed by Korshunov et al.~\cite{FantasyID}. FantasyID contains ID cards with diverse design styles and languages, featuring only the faces of real people. To simulate a realistic KYC scenario, the cards from FantasyID were printed and captured with three different devices, constituting the bona fide class. The authors have emulated digital forgery/injection attacks that a malicious actor could perform to tamper with the IDs using the existing generative tools

Tapia et al.~\cite{tapia_first_2024, tapia_second_2025} demonstrated, in the last two PAD on ID Cards competitions, the challenge of achieving generalization across multiple countries due to the lack of primary datasets containing a large number of bona fide images. 
Regarding foundation models, Tapia et al.~\cite{tapia2025FM} proposed a PAD system using CLIP~\cite{CLIP} and DinoV2~\cite{Dinov2} to create two downstream models that fuse the best embeddings from each one. 

A related case was developed by Mu{ñ}oz-Haro et al.~\cite{FakeIDet}, where a PAD dataset on an ID Card system was created. The authors focusing on two levels of anonymization for an ID (i.e., fully- and pseudo-anonymized), and different patch size configurations, varying the amount of sensitive data visible in the patch image. This dataset has 48,400 patches from bona fide and fake IDs, taken from 30 subjects in total.

\section{Datasets}
\label{sec:dataset}
The datasets comprise ID Cards from five countries: Chile, Mexico, and Poland, Portugal, and Spain. There are four distinct categories of data: {\it Bona Fide, Border, Screen, and Printed}. The {\it Border, Screen}, and {\it Printed} categories correspond to different types of attacks. All the datasets are imbalanced, i.e., attack samples outnumber the number of bona fide samples. 

The bona fide image refers to a  genuine ID Card/Passport issued by a government agency, which is presented to the capture device. The printed attack is a photographic reproduction of the bona fide image. The Border attack image is a forgery that can be digitally or manually modified to include certain ID Card swap areas/faces. The screen attack displays a photograph of the ID card, which is the screen-capture version of the bona fide image.

\begin{figure*}[]
\centering
\includegraphics[scale=0.14]{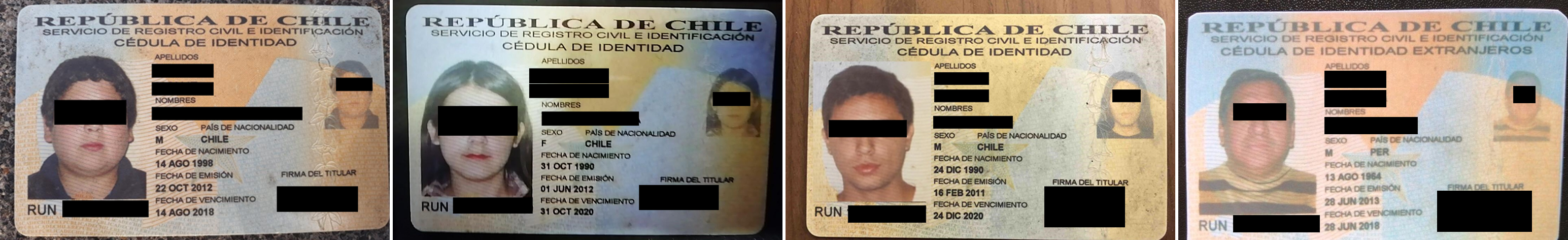}
\includegraphics[scale=0.14]{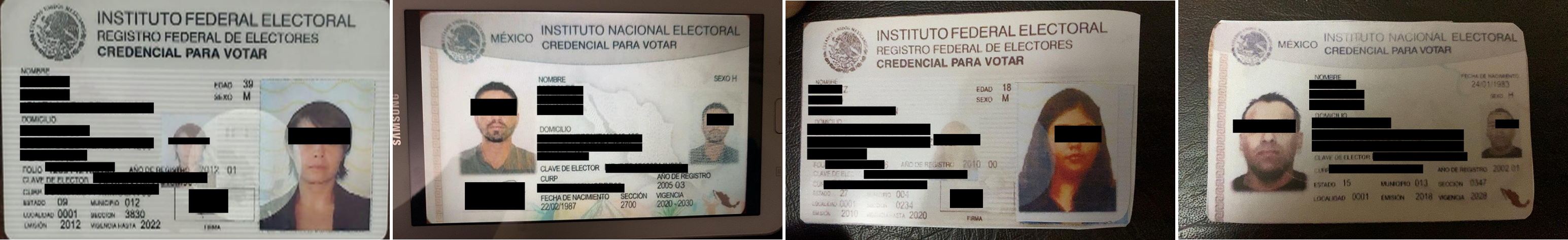}
\caption{Examples of four different attack types. From top to bottom: Chile and Mexico ID Card datasets. From left to right: Bona Fide, Screen attack, Border attack. and Printed attack.}
\label{fig:real}
\end{figure*}

The Chile and Mexico datasets include all four categories, and the data comes from genuine ID Cards. Examples of each category are presented in Figure \ref{fig:real}. 
The data quantity distributions are, for Chile, train: 24,603, test: 8,174, validation: 8,173, for Mexico, train: 2,275, test: 763, validation: 758. The Chile dataset has the largest quantity of data. Its training set is used to train the model, the evaluation set is used for validation during training, and the test set, along with the datasets from other countries, is used for the final evaluation.

\begin{figure*}[]
\centering
\includegraphics[scale=0.16]{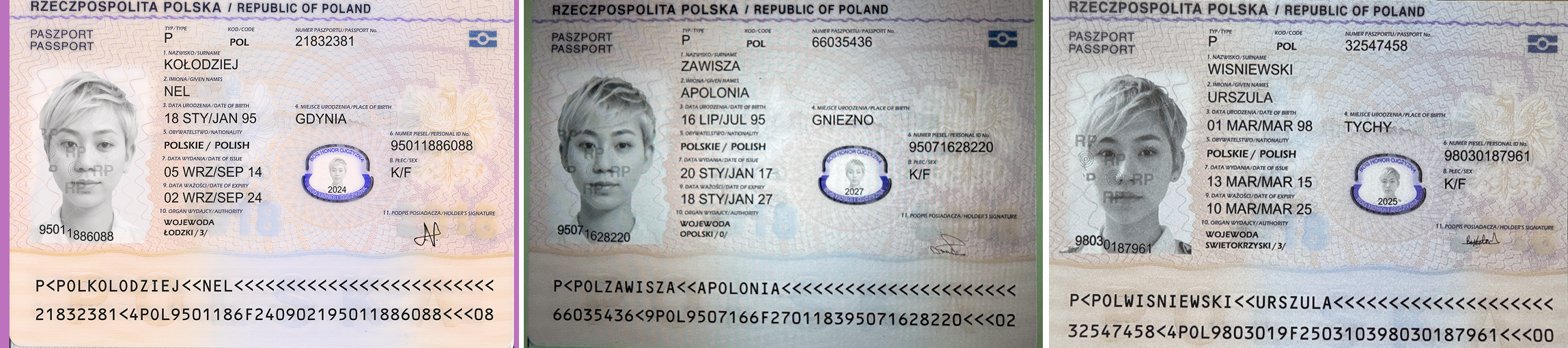}
\includegraphics[scale=0.16]{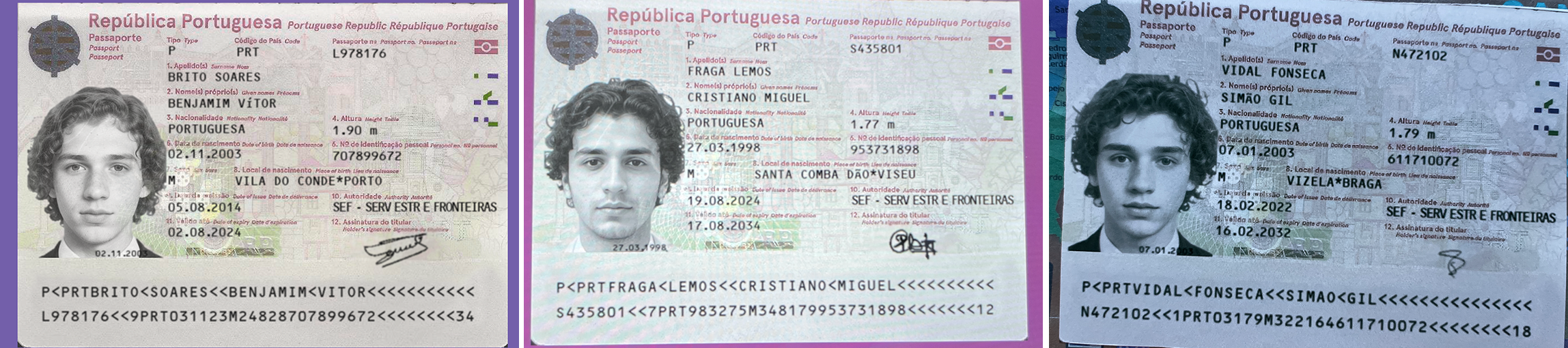}
\includegraphics[scale=0.16]{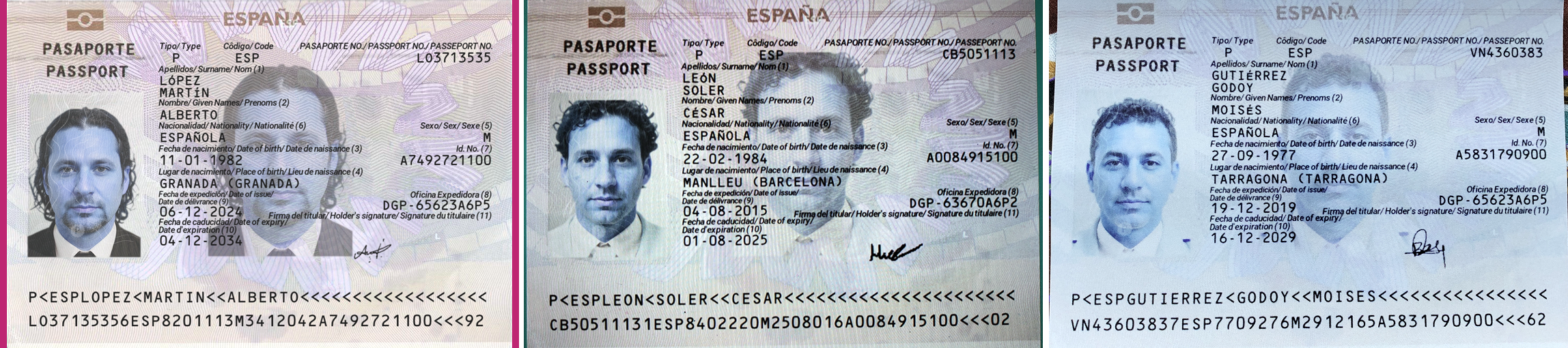}
\caption{Examples of four different attack types. From top to bottom: Poland, Portugal, and Spain ID Card datasets. From left to right: Bona Fide, Screen attack, and Printed attack.}
\label{fig:synthetic}
\end{figure*}

The datasets of Poland, Portugal, and Spain are from open-access synthetic passport datasets ~\cite{tapia2025synid}. They only include simulated-synthetic {\it Bona fide}, and its {\it Screen}, and {\it Printed} categories, manually created. Examples of each category are presented in Figure~\ref{fig:synthetic}. Each category contains approximately 1,000 samples. All of them will be used as the test set in a cross-dataset scenario.

% real ID card example

% Synthetic ID card example

\section{Method}
\label{sec:Method}
To better demonstrate the effectiveness of multimodal model architectures for PAD on ID Cards, we develop a three-model evaluation framework comprising a classical deep learning model, a unimodal model, and a compact multimodal model. 
The deep learning and unimodal models are treated as baseline methods, while the multimodal models serve as ablation controls for one another. 

To ensure consistent performance computation as defined in Section \ref{sec:metrics}, all models are constrained to output a two-class probability score for each ID Card. For all discriminative models, including the deep learning baseline, the unimodal model, and the multimodal model employing a discriminative structure, the bona fide class is mapped to label 0, and the attack class is mapped to label 1.

\subsection{Deep Learning}
\label{method_deep_learning}
Based on the strong and consistent performance reported in prior work, DenseNet-121 \cite{huang2017densely, mudgalgundurao2022pixel} was selected as the baseline model. The backbone was pretrained on ImageNet \cite{deng2009imagenet} to evaluate the performance of conventional convolutional architectures for ID Card PAD. The network’s densely connected structure benefits efficient feature reuse and stable optimization under limited data. All input images are resized to $384\times384$ pixels, which are linearly projected into $1,024$ dimensional embeddings

To adapt the network to binary PAD classification, the original classification layer is replaced with a two-unit output head corresponding to the bona fide and attack classes, optionally regularized with dropout. This model serves as a baseline for comparison with the multimodal approaches.

\subsection{Unimodal Model}
\label{Unimodal_model}
In this experiment, we focus on the visual modality in a unimodal setting by adopting SigLIP \cite{zhai2023sigmoid}, a CLIP-style vision–language architecture that replaces the conventional softmax-based contrastive loss with a sigmoid-based pairwise ranking objective. We use the SigLIP-SO400M-Patch14-384 \footnote{\url{https://huggingface.co/google/siglip-so400m-patch14-384}} variant and retain only its Vision Transformer (ViT) component \cite{dosovitskiy2020image}, thereby isolating the contribution of high-capacity visual representations without multimodal fusion. This configuration serves as a vision-only ablation, allowing us to quantify the effect of removing language and cross-modal interaction mechanisms present in multimodal PAD models.

All input images are resized to $384 \times 384$ pixels and partitioned into non-overlapping $14 \times 14$ patches, which are linearly projected into $1,152$ dimensional embeddings. The ViT backbone consists of $27$ Transformer layers \cite{vaswani2017attention,dosovitskiy2020image}, each comprising multi-head self-attention and feed-forward sublayers with residual connections and layer normalization, using GELU activations \cite{hendrycks2016gaussian}.
After the final Transformer layer, patch-level representations are aggregated via global mean pooling to produce a single image-level embedding $h \in \mathbb{R}^{1152}$.

Instead of fusing the visual embedding with language representations, as in multimodal architectures, the vision-level embedding is passed through a single linear projection layer to obtain a scalar logit, which is subsequently transformed into an attack probability using a sigmoid function. The model is trained using binary cross-entropy loss and functions as a unimodal ablation baseline for comparison against multimodal PAD architectures.

\subsection{Multimodal}
\label{Multimodal}
In the multimodal setting, our analysis concentrates primarily on the textual and visual modalities, while also accounting for practical deployment constraints. We adapt the SmolVLM2 architecture \cite{marafioti2025smolvlm} into a binary PAD classifier for ID Cards. SmolVLM2 is a compact vision–language model that couples a shape-optimized SigLIP Vision Transformer (ViT) backbone with the SmolLM2 language model through a shared multimodal embedding space. The model is evaluated under both zero-shot inference (frozen backbone) and fine-tuned settings. Across both settings, training and evaluation are conducted using a fixed task prompt to ensure consistent conditioning and fair comparison, which is shown in Table \ref{tab:prompts}.

\begin{table*}[t]
\centering
\caption{Prompt template for the multimodal PAD classifier.}
\label{tab:prompts}
\begin{tabular}{@{}p{2.2cm}p{13.3cm}@{}}
\toprule
\textbf{Role} & \textbf{Content} \\
\midrule
\textbf{User} &
\begin{minipage}[t]{\linewidth}
\raggedright
\textbf{ROLE:} You are a senior forensic document examiner specialising in ID Card photos (camera-captured images under varied backgrounds).\\[0.5em]

\textbf{TASK:} You will receive exactly one ID Card image. Perform presentation attack detection (PAD) and determine its authenticity.\\[0.5em]

\textbf{ATTACK CHECKLIST:}
\begin{itemize}
    \item \textbf{A: Print} — A reproduction of an ID Card that has been re-captured by camera or scanner.
    \item \textbf{B: Border} — Local region replacement (especially the portrait) that has been pasted, digitally swapped, or altered, leaving visible seams, misalignments, or texture inconsistencies.
    \item \textbf{C: Screen} — The ID or its photo displayed on an electronic device (phone, tablet, monitor) and re-photographed.
    \item \textbf{D: Bonafide\footnote{For prompts, the \textit{bona fide} word is described as a single word (\textit{BONAFIDE})}} — Otherwise, the image is considered bona fide.
\end{itemize}

\textbf{DECISION POLICY:}
\begin{itemize}
    \item Base your judgment strictly on visible forensic evidence; do not assume bona fide by default.
    \item At most one attack type should be assigned per sample.
    \item Isolated minor flaws are neutral and not sufficient evidence of an attack.
\end{itemize}

\textbf{CONSTRAINTS:}
\begin{itemize}
    \item If evidence remains ambiguous, choose the class with the higher calibrated likelihood (never default unconditionally to bona fide).
\end{itemize}

\textbf{FINAL TASK:} Analyze the provided ID Card image, apply the Attack Checklist, and output an authenticity label.\\[0.3em]
Your output is strictly limited and can only be attack/bona fide.
\end{minipage}
\\
\bottomrule
\end{tabular}
\end{table*}

In the zero-shot setting, PAD is formulated as a generative classification problem using the model’s native next-token prediction mechanism, without updating any model parameters. Given an input ID-card image and a fixed task prompt, the image is encoded by the ViT backbone and jointly processed with the textual prompt through the language model using the standard multimodal chat template. A single forward pass over the image–prompt context is performed to obtain the next-token probability distribution, together with the corresponding key–value cache.

\begin{figure*}[]
\centering
\includegraphics[scale=0.35]{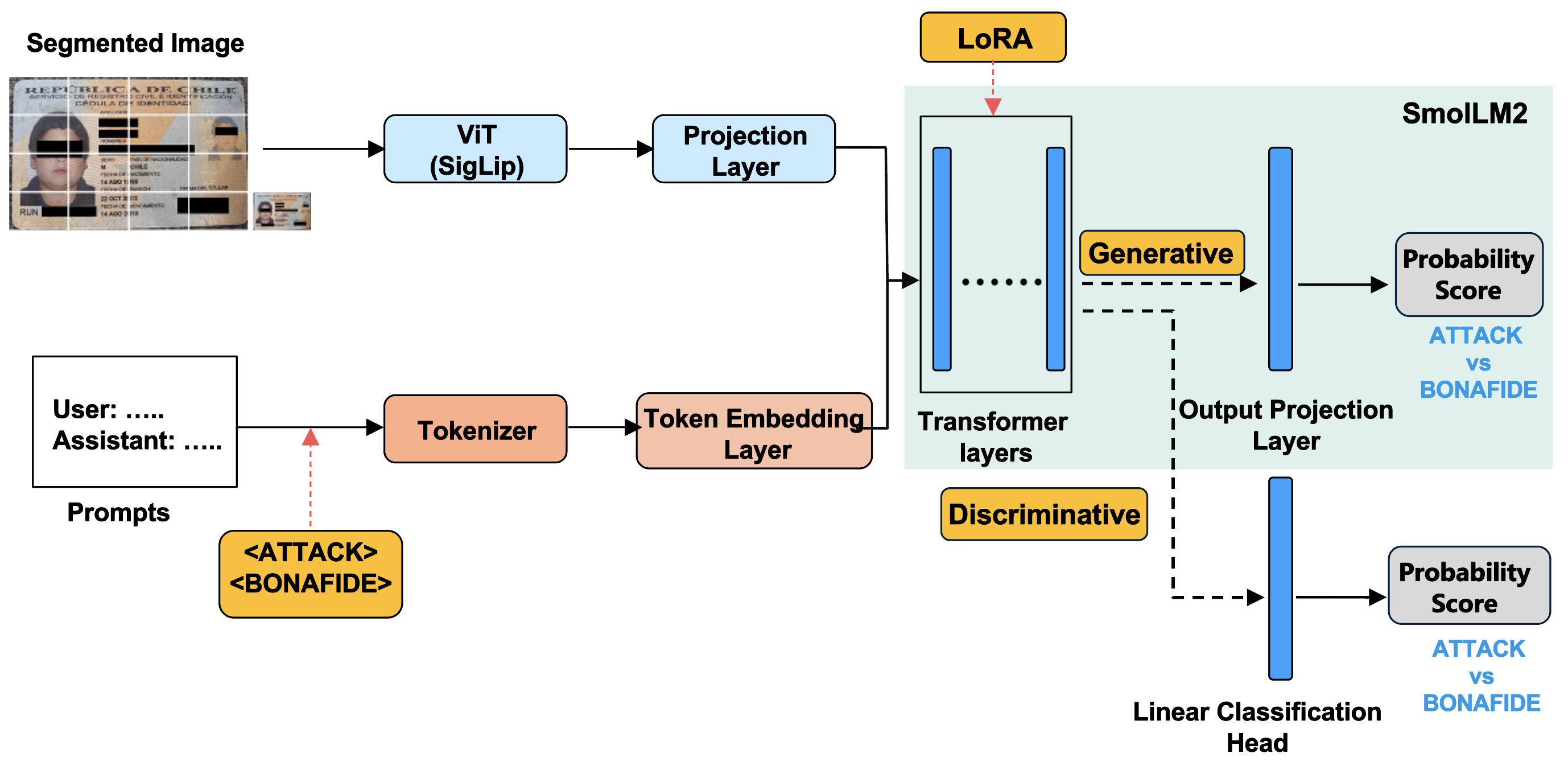}
\caption{The two different structures of SmolVLM2 in PAD on ID Cards. special token \texttt{<FAKE>} and \texttt{<BONAFIDE>} only add in Generative structure. LoRA only adds in the fine-tuning process.}
\label{fig:Structure_SmolVLM2}
\end{figure*}

Binary classification is performed by computing the normalized likelihoods of the Attack and Bona fide labels conditioned on the image–prompt context. These likelihoods are normalized using a softmax function to produce the final probability scores for PAD decision-making.\\
To fine-tune the SmolVlm2, we design two new alternative modelling structures, \textit{Generative} and \textit{Discriminative}, each achieving binary classification through distinct mechanisms. The two structures are illustrated in Figure \ref{fig:Structure_SmolVLM2}. 

In the \textit{generative structure}, SmolVLM2 is fine-tuned to perform PAD as a constrained sequence generation task, where the model is restricted to generating predefined label tokens (bona fide or attack) conditioned on the input ID Card image and prompt, and probability scores are derived from the corresponding generation likelihoods.

In contrast, the \textit{discriminative structure} directly produces two class-specific scores associated with bona fide and attack through a classification-oriented training objective, without explicit sequence generation.

In addition, to assess the tradeoff between parameter scale and the capacity of multimodal models to perform PAD on ID Cards, we adopt two differently sized SmolVLM2 variants: \textbf{SmolVLM2-500M-Video-Instruct}\footnote{\url{https://huggingface.co/HuggingFaceTB/SmolVLM2-500M-Video-Instruct}} and \textbf{SmolVLM2-2.2B-Instruct}\footnote{\url{https://huggingface.co/HuggingFaceTB/SmolVLM2-2.2B-Instruct}}, comprising approximately 500 million and 2.2 billion parameters, respectively. 

\paragraph{Generative Structure}
The input to the model consists of a single ID Card image paired with a fixed prompt. In the multimodal generative setting, SmolVLM2 processes visual and textual inputs jointly to perform presentation attack detection.
On the visual side, to align with the structural requirements of SmolVLM2, the \textit{AutoProcessor} (provided by the transformers library) \footnote{\url{https://huggingface.co/docs/transformers/en/main_classes/processors}} applies the following modifications. Each image is first read as an RGB matrix and uniformly scaled so that its longer side equals $1,536$ pixels. Zero-padding is applied to the image boundaries until both width and height are divisible by $384$. The padded image is then evenly partitioned into multiple $384 \times 384$ image tiles and a downscaled global copy. After this process, the visual tensor is represented as $\left[B, N, 3, 384, 384\right]$, where $B$ denotes the batch size and $N$ the number of image tiles per sample.

Each tile of shape $[3,\,384,\,384]$ is processed by a SigLIP-style ViT. The encoder first 
divides the tile into a regular grid of non-overlapping patches of 
size $P \times P$, resulting in a patch grid of 
$H \times W$ patches, where
\[
H = 384 / P,\qquad 
W = 384 / P.
\]

Each patch is linearly projected into a $C$-dimensional embedding 
vector by the ViT patch-embedding layer, producing a sequence of 
visual tokens. The resulting representation has the shape:
\begin{equation}
[B,\,N,\,HW,\,C],
\end{equation}
where $H, W$ is the length of the patch-token sequence, and $C$ denotes 
the ViT hidden dimension.

To reduce the number of visual tokens while preserving
local structure, SmolVLM2 applies a pixel-shuffle (space-to-depth) 
token compression operation to each tile’s feature map. Given a 
compression factor $r$, this operation decreases the spatial 
resolution from $(HW)$ to $(H/rW/r)$ while folding the 
corresponding spatial information into the channel dimension, yielding:
\begin{equation}
[B,\,N,\,C \cdot r^{2},\,H/rW/r].
\end{equation}
The compressed feature maps are then flattened along their spatial 
dimensions to form $N$ tile-level visual token groups, each containing 
$(H/r)(W/r)$ tokens. Concatenating all tile-level tokens results in the 
final visual token sequence:

\begin{equation}
[B,\,N \cdot L,\,C'],
\end{equation}
where $L = (H/r)(W/r), C' = C r^{2}$ denotes the effective embedding dimension after 
pixel shuffling. 

These visual tokens are subsequently projected by a learned projection layer into the shared multimodal embedding space of dimension $D$, corresponding to the hidden size of the SmolLM2 Transformer. In \textbf{SmolVLM2-500M-Video-Instruct}, $D$ equals $960$, and in \textbf{SmolVLM2-2.2B-Instruct}, $ D$ equals $2048$, respectively, and merge with the textual token embedding.

The prompt is formulated in SmolVLM2’s dialogue template, ensuring the input format aligns with the interaction structure used by the model during pre-training. A special marker indicating the next token as the assistant’s response is added to the end of the template, indicating that the following tokens should be generated as the assistant’s response. This marker can be added automatically by using the Hugging Face method
\textit{apply\_chat\_template} \footnote{\url{https://huggingface.co/docs/transformers/en/chat_templating}}.

The prompt text is then converted into a series of discrete identifiers (token IDs) by a tokenizer.  To transform the problem from natural language generation into binary classification, two special label tokens \texttt{<BONAFIDE>} and \texttt{<ATTACK>} are added to the tokenizer vocabulary and update independent learnable embeddings. During training, only the ground-truth label token is appended to the end of the input sequence and treated as the supervised target in the loss calculation. This setup guides the model to generate the correct label token, either \texttt{<BONAFIDE>} or \texttt{<ATTACK>} as the output of the classification task.

During this process, three tensors are built: input\_ids and attention\_mask are generated by the tokenizer according to the dialogue template, and the label mask is designed specifically for this task to constrain the loss to the final output token. 

\begin{itemize}
  \item \texttt{input\_ids}: The most basic input tensor of a language model. It records the index number of each token in the complete text sequence, corresponding to a unique position in the vocabulary. For a sample, the length of input\_ids is equal to the number of tokens in the input sequence, including task prompt text, special symbols (such as dialogue start and end markers), and additional category labels \texttt{<BONAFIDE>} or \texttt{<ATTACK>}.
  \item \texttt{attention\_mask}: A binary vector of the same length as `input\_ids` indicates which positions participate in the attention calculation. Positions with a value of 1 represent true tokens that should be paid attention to by the model; positions with a value of 0 represent padding or invalid parts that do not participate in any self-attention or gradient propagation.
  \item \texttt{labels}: It defines the supervision signal for the training phase. It has the same shape as `input\_ids`, but its function is completely different: except for the last position, all other positions are padded with -100, indicating that these positions are ignored when calculating the cross-entropy loss. Only the last position corresponds to the index of the true class token of the current sample, i.e., the index value of \texttt{<BONAFIDE>} or \texttt{<ATTACK>}
\end{itemize}

When multiple samples are input in a batch, padding is needed on the right side to ensure consistent sequence lengths. The padded input\_ids are appended with $0s$, the attention\_mask is set to $0$ after padding, while the labels remain $-100$ at their respective positions.

After tokenization, each token ID is mapped to its vector representation through an embedding lookup
parameterized by the embedding matrix $E \in \mathbb{R}^{|V| \times D}$, where $|V|$ denotes the vocabulary
size and $D$ is the hidden dimension of the SmolLM2 Transformer (also shared with the visual projection
dimension). For a given batch, this produces the initial text embedding tensor
$\mathbf{H}_0 \in \mathbb{R}^{B \times T^* \times D}$, where $T^*$ is the padded sequence length.
This embedding operation transforms discrete token indices into continuous $D$-dimensional vectors.

The final multimodal input sequence is constructed by directly concatenating the visual and textual embeddings along the sequence dimension, resulting in a unified tensor of shape $\mathbb{R}^{B \times (T + N\!\cdot\!L) \times D}$. The SmolLM2 Transformer then processes this sequence through cross-modal self-attention, producing a set of hidden representations $\mathbf{H} \in \mathbb{R}^{B \times T \times D}$ corresponding to the textual positions.
To convert these hidden states into vocabulary logits, SmolLM2 applies the output projection layer, implemented as a linear mapping parameterized by the embedding matrix $E \in \mathbb{R}^{|V|\times D}$. Specifically, the logits are computed as
\[
\mathbf{Z} = \mathbf{H} E^\top \in \mathbb{R}^{B \times T \times |V|}.
\]
Each element $Z_{b,t,v}$ represents the unnormalized score (logit) assigned to token $v$ at position $t$ in sample $b$. Applying a softmax function along the vocabulary dimension yields the conditional token probabilities, where v' ranges over the entire vocabulary:
\[
P(v_t \mid \mathbf{v}_{<t}) = 
\frac{\exp(Z_{b,t,v})}{\sum_{v'} \exp(Z_{b,t,v'})}.
\]
For binary generative classification, the model’s output is obtained by reading the probabilities associated with the label tokens \texttt{<BONAFIDE>} and \texttt{<ATTACK>}.

\paragraph{Discriminative Structure}
This block largely preserves the visual encoding and multimodal fusion of the generative structure; the key difference lies in how supervision is expressed. Instead of appending explicit label tokens (\texttt{<ATTACK>} / \texttt{<BONAFIDE>}) to the prompt and modeling their generation probability, the discriminative variant removes label tokens entirely. A single positional marker (\texttt{mark\_labels}) is inserted to indicate the location where the model is expected to form its authenticity judgment, while all other positions are masked with $-100$. This marker is not a learnable label token; it simply designates the textual position whose hidden representation will be used for classification.

At the model level, the SmolLM2 backbone remains unchanged, while a lightweight classification head operates on the final-layer hidden representation associated with the \texttt{mark\_labels} token. This hidden vector, $\mathbf{h}\in\mathbb{R}^D$, already integrates the combined
visual and textual context through SmolLM2's cross-modal self-attention. The classification head is a small feed-forward module that operates only on this single vector. Concretely, the head
consists of:
\[
\mathbf{h}' = \mathrm{Dropout}(\mathbf{h}),
\qquad
z = \mathbf{w}^{\top}\mathbf{h}' + b,
\]
where $\mathbf{w}\in\mathbb{R}^D$ and $b\in\mathbb{R}$. This produces a scalar logit
$z$ representing evidence for the ``Attack'' class.

Unlike the generative model, which implicitly encodes supervision
via the likelihood of generating \texttt{<ATTACK>} or \texttt{<BONAFIDE>}
through the vocabulary projection, the discriminative head avoids
the $|V|$-dimensional output projection entirely. Instead, the scalar
$z$ is paired with a fixed zero logit to construct a two-dimensional
logit vector \([z,\,0]\) corresponding to
\([\,\text{Attack},\,\text{Bona\_Fide}\,]\). Applying a softmax over
these two values yields the probability scores of Attack and Bona Fide.

\section{Metric}
\label{sec:metrics}
This section describes the metrics used to evaluate image generation and PAD ID Card performance. The international standard ISO/IEC 30107-3\footnote{\url{https://www.iso.org/standard/67381.html}} presents methodologies for evaluating the detection performance of PAD algorithms for biometric systems. The Attack Presentation Classification Error Rate (APCER) measures the proportion of attack presentations for each Presentation Attack Instrument (PAI) incorrectly classified as bona fide presentations. This metric is calculated for each PAI, considering the worst-case scenario. Equation~\ref{eq:apcer} presents the computation of the APCER metric, in which the value of $N_{PAIS}$ corresponds to the number of attack presentation, where $RES_{i}$ for the $i$th image is $1$ if the algorithm classifies it as an attack presentation, or $0$ if it is classified as a bona fide presentation (real image).

\begin{equation}\label{eq:apcer}
    {APCER_{PAIS}}=1 - (\frac{1}{N_{PAIS}})\sum_{i=1}^{N_{PAIS}}RES_{i}
\end{equation}

Additionally, the Bona fide Presentation Classification Error Rate (BPCER) metric measures the proportion of bona fide presentations incorrectly classified as attack presentations. The BPCER metric is formulated according to equation~\ref{eq:bpcer}, where $N_{BF}$ corresponds to the number of bona fide presentation images, and $RES_{i}$ takes identical values to those of the APCER metric. Furthermore, the Equal Error Rate (EER), which is the value when the APCER equals the BPCER, is also reported.
\begin{equation}\label{eq:bpcer}
    BPCER=\frac{\sum_{i=1}^{N_{BF}}RES_{i}}{N_{BF}}
\end{equation}

These metrics effectively measure the degree to which the algorithm confuses presentations of attack images with bona fide images and vice versa.

\section{Experiment and Results}
\label{sec:EAR}

\begin{table*}[t]
\centering
\caption{PAD performance (EER and BPCER) of DenseNet, SigLIP-SO400M, and SmolVLM2 (500M and 2.2B) under zero-shot, discriminative, and generative settings on realistic ID Card datasets from Chile and Mexico, where B10, B20, and B100 denote BPCER10, BPCER20, and BPCER100, respectively.}
\setlength{\tabcolsep}{5pt}
\renewcommand{\arraystretch}{1.2} 

\begin{tabular}{lcccccccc}
\toprule
 & \multicolumn{4}{c}{\textbf{Chile}} 
 & \multicolumn{4}{c}{\textbf{Mexico}} \\
\cmidrule(lr){2-5}\cmidrule(lr){6-9}

\textbf{Models} &
\textbf{EER} & \textbf{B10} & \textbf{B20} & \textbf{B100} &
\textbf{EER} & \textbf{B10} & \textbf{B20} & \textbf{B100} \\
\midrule

DenseNet
& 2.21 & 0.51 & 0.68 & 7.77
& 36.77 & 63.47 & 74.95 & 88.82 \\

SigLIP-SO400M (Unimodal)
& \mycc0.85 & \mycc0.00 & \mycc0.00 & \mycc0.43
& \mycc9.17 & \mycc8.38 & \mycc15.37 & \mycc34.43 \\
\midrule

SmolVLM2-500M-Zero-Shot
& 50.46 & 90.78 & 95.30 & 99.15
& 41.78 & 74.85 & 84.93 & 97.11 \\

SmolVLM2-2.2B-Zero-Shot
& 45.26 & 85.22 & 91.12 & 97.36
& 22.39 & 39.80 & 54.23 & 90.55 \\
\midrule

SmolVLM2-500M-Gen.
& 1.96 & 0.17 & 0.68 & 6.23
& 25.88 & 56.86 & 66.61 & 89.52 \\

SmolVLM2-2.2B-Gen.
& \mycc0.93 & \mycc0.34 & \mycc0.34 & \mycc0.76
& \mycc5.99 & \mycc4.24 & \mycc7.48 & \mycc16.46 \\
\midrule

SmolVLM2-500M-Disc.
& 25.02& 56.70 & 74.85 & 91.63
& 39.79 & 64.34 & 72.82 & 81.05 \\

SmolVLM2-2.2B-Disc.
& 17.08 & 30.03 & 45.69 & 77.71
& 19.29 & 35.66 & 46.63 & 64.59 \\
\bottomrule
\end{tabular}

\label{tab:pad_Chile_Mexico}
\end{table*}

\begin{table*}[t]
\centering
\caption{PAD performance (EER and BPCER) of DenseNet, SigLIP-SO400M, and SmolVLM2 (500M and 2.2B) under zero-shot, discriminative, and generative settings on Poland, Portugal and Spain, where B10, B20, and B100 denote BPCER10, BPCER20, and BPCER100, respectively.}
\setlength{\tabcolsep}{5pt}
\renewcommand{\arraystretch}{1.2}

\begin{tabular}{lcccccccccccc}
\toprule
 & \multicolumn{4}{c}{\textbf{Poland}} 
 & \multicolumn{4}{c}{\textbf{Portugal}} 
 & \multicolumn{4}{c}{\textbf{Spain}} \\
\cmidrule(lr){2-5}\cmidrule(lr){6-9}\cmidrule(lr){10-13}

\textbf{Models} &
\textbf{EER} & \textbf{B10} & \textbf{B20} & \textbf{B100} &
\textbf{EER} & \textbf{B10} & \textbf{B20} & \textbf{B100} &
\textbf{EER} & \textbf{B10} & \textbf{B20} & \textbf{B100} \\
\midrule

DenseNet
& 23.58 & 48.97 & 63.35 & 84.63
& 41.52 & 69.56 & 83.05 & 95.86
& 36.38 & 63.39 & 75.76 & 94.98 \\

SigLIP-SO400M(Unimodal)
& \mycc20.97 & \mycc35.76 & \mycc53.10 & \mycc88.57
& 19.31 & 74.19 & 93.30 & 99.90
& 49.04 & 100.00 & 100.00 & 100.00 \\
\midrule

SmolVLM2-500M-Zero-Shot
& 61.25 & 98.32 & 99.51 & 100.00
& 49.70 & 95.47 & 98.23 & 99.51
& 44.82 & 84.24 & 93.20 & 98.82 \\

SmolVLM2-2.2B-Zero-Shot
& 51.21 & 96.06 & 99.11 & 99.90
& \mycc17.60 & \mycc38.03 & \mycc65.71 & \mycc93.50
& \mycc16.15 & \mycc26.80 & \mycc41.18 & \mycc73.99\\
\midrule

SmolVLM2-500M-Gen.
& 26.98 & 95.99 & 99.80 & 100.00
& 75.69 & 100.00 & 100.00 & 100.00
& 88.76 & 100.00 & 100.00 & 100.00 \\

SmolVLM2-2.2B-Gen.
& 24.82 & 69.63 & 85.23 & 96.72
& 28.82 & 99.33 & 99.90 & 99.90
& 30.62 & 99.80 & 99.90 & 100.00 \\
\midrule

SmolVLM2-500M-Disc.
& 24.46 & 62.95 & 80.04 & 94.97
& \mycc9.66 & \mycc8.87 & \mycc22.84 & \mycc53.69
& 35.49 & 82.07 & 89.37 & 97.40 \\

SmolVLM2-2.2B-Disc.
& 32.57 & 83.39 & 93.89 & 99.11
& 20.02 & 71.15 & 87.15 & 99.31
& 32.93 & 82.59 & 92.37 & 99.21 \\
\bottomrule
\end{tabular}
\label{tab:pad_Poland_Portugal_Spain}
\end{table*}

\begin{figure*}[]
\centering
\caption{Cross-Country DET Curves for Presentation Attack Detection Using DenseNet, SigLIP-SO400M, and SmolVLM2 (500M and 2.2B) under Zero-Shot, Discriminative, and Generative Settings considering Chile, Mexico, Poland, Portugal and Spain.}
\includegraphics[scale=0.08]{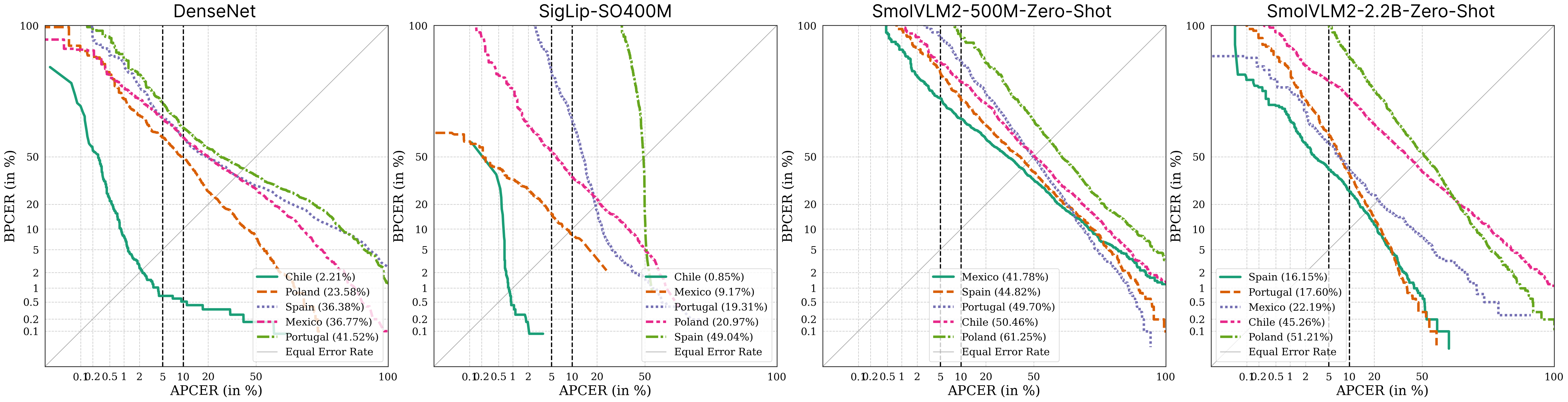}
\includegraphics[scale=0.08]{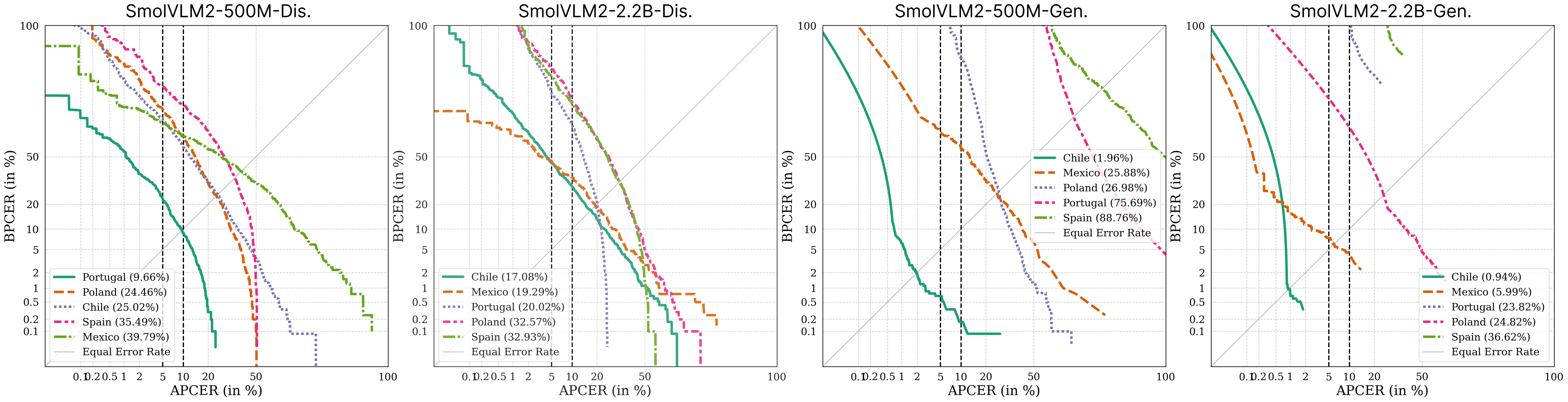}
\label{fig:real-idcards-examples}
\end{figure*}

\subsection{Experiment 1 - Deep Learning}
Training is performed using a class-weighted cross-entropy loss with optional label smoothing, where class weights are computed from the inverse class frequencies in the training set to address imbalance between bona fide and attack samples. Model parameters are optimized using the AdamW optimizer \cite{loshchilov2017decoupled} with a cosine learning-rate schedule \cite{loshchilov2016sgdr} and linear warm-up to ensure stable convergence \cite{kalra2024warmup}.
Key hyperparameters, including input batch size, learning rate, weight decay, classifier dropout rate, data augmentation strength, and warm-up duration, are automatically selected using the Optuna tool \cite{akiba2019optuna}. During hyperparameter search, models are trained for 10 epochs and evaluated based on validation loss. After hyperparameter selection, the model is retrained from scratch using the best configuration for up to 20 epochs. 

Results on genuine ID Card datasets (Chile and Mexico) are reported in Table \ref{tab:pad_Chile_Mexico}. On the Chile dataset, DenseNet achieves a low EER of 2.21\%. However, performance degrades substantially in Mexico, where the EER increases to 36.77\%, and BPCER remains high across all operating points. 

Performance on synthetic ID Card datasets (Poland, Portugal, and Spain) is summarized in Table \ref{tab:pad_Poland_Portugal_Spain}. Across all synthetic datasets, DenseNet exhibits consistently high EER values exceeding 23\%, with BPCER increasing sharply at stricter operating points. Compared to genuine datasets, the model shows a remarkable loss of discriminative capability.

\subsection{Experiment 2 - Unimodal Model}
Training is conducted using the Binary Cross-Entropy loss \cite{bishop2006pattern}, which measures the discrepancy between the predicted attack probability and the ground-truth label $y \in {0,1}$. The loss is defined as:
\begin{equation}
\mathcal{L} =
[y_i \log P{\text{Attack}}+ (1 - y_i) \log (1 - P_{\text{Attack}})],
\end{equation}
where $y$ denotes the corresponding ground-truth label.
Optimization is performed using the AdamW optimizer with a weight decay coefficient of $0.01$ \cite{ng2004feature}. To ensure stable training, three learning rates ($1\times10^{-4}$, $3\times10^{-5}$, and $1\times10^{-5}$) are evaluated.  A cosine annealing learning-rate scheduler is applied at the step level. The model is trained for 14 epochs with a mini-batch size of 16.

Light data augmentation is applied during training in the form of small random image rotations, while validation and test samples are evaluated without augmentation. Each configuration is trained on the Chile training set and evaluated on the Chile validation set. Once the optimal learning rate is determined on the Chile dataset, the same model configuration is evaluated on the remaining genuine and synthetic ID Card datasets to assess cross-dataset generalization.

Results on genuine ID Card datasets are reported in Table \ref{tab:pad_Chile_Mexico}. On the Chile dataset, the unimodal model achieves strong performance, with an EER of 0.85\% and zero-shot BPCER at both BPCER10 and BPCER20, indicating reliable separation between bona fide and attack samples under similar acquisition conditions. In Mexico, performance degrades to an EER of 9.17\%, with BPCER increasing slightly at stricter operating points. 

Performance on synthetic ID Card datasets is summarized in Table \ref{tab:pad_Poland_Portugal_Spain}. Compared with the deep learning baseline, the unimodal model shows mixed behavior across synthetic datasets. For the Poland and Portugal, SigLIP-SO400M achieves EER values that are comparable to or lower than those of DenseNet. However, the results for the Spain dataset constitute a clear outlier, with markedly higher EER and BPCER values that reach extreme levels at stricter operating points.

\subsection{Experiment 3 - Multimodal Model}
\paragraph{Zero-Shot}
In the zero-shot setting, all model parameters remain frozen, and fine-tuning is performed. Evaluation on the Chilean test set and on datasets from other countries is performed using a single, fixed discrimination prompt that is applied consistently across all datasets. For each test sample, the model outputs posterior probabilities for the Attack and Bona Fide classes conditioned on the image–prompt input. Class probabilities are obtained directly from the normalized likelihoods of the target labels.

Results on genuine ID Card datasets are reported in Table \ref{tab:pad_Chile_Mexico}. In Chile, zero-shot SmolVLM2 exhibits very high error rates, with EER exceeding 45\% for both the 500M and 2.2B variants. Similar behavior is observed in Mexico dataset, where EER remains above 20\% even for the larger model. BPCER values are correspondingly high across all operating points.

Performance on synthetic ID Card datasets is summarized in Table \ref{tab:pad_Poland_Portugal_Spain}. Across all synthetic datasets, zero-shot performance remains consistently weak, with EER values typically above 45\% for the 500M model and above 15–20\% for the 2.2B model. BPCER values increase sharply at stricter operating points, indicating unstable decision boundaries under domain-shifted conditions.

\paragraph{Generative Structure}
Fine-tuning is performed using Low-Rank Adaptation (LoRA) \cite{hu2022lora} applied exclusively to the text decoder backbone component of SmolVLM2, while the vision encoder remains frozen. LoRA parameterizes the trainable weight update as a low-rank matrix
\begin{equation}
\Delta W = \frac{\alpha}{r} \, B A ,
\end{equation}
where $A \in \mathbb{R}^{r \times d_{\text{in}}}$ and $B \in \mathbb{R}^{d_{\text{out}} \times r}$ are trainable matrices, $r \ll \min(d_{\text{in}}, d_{\text{out}})$ denotes the rank, and $\alpha$ is a scaling factor. During fine-tuning, only $A$ and $B$ are optimized, while all original model parameters remain frozen. At initialization, $B$ is set to zero and $A$ is initialized with small random values, such that $W' \approx W$; during fine-tuning, only the LoRA parameters receive gradients. In all experiments, LoRA is applied exclusively to the language-side projection layers of the SmolLM2 backbone, including Query Projection, Key Projection, Value Projection, and Output Projection in the self-attention modules, as well as Up Projection, Down Projection, and Gate Projection in the feed-forward networks. 

A fixed discrimination prompt is used consistently across all datasets, and two masks are applied during training: a padding mask to prevent attention to padded positions introduced during sequence alignment, and a label mask that restricts supervision to the final label token only. As a result, the cross-entropy loss is computed exclusively on the supervised label position,
%The generative PAD objective is formulated as a next-token prediction task, where the model is trained to generate a single label token corresponding to either \texttt{<ATTACK>} or \texttt{<BONAFIDE>} conditioned on the image--prompt input. 
\[
\mathcal{L}
= - \log P\!\left(t_{T+1}\mid t_{\le T}\right),
\qquad
t_{T+1}\in\mathcal{Y},
\]
where $\mathcal{Y}=\{\texttt{<ATTACK>},\texttt{<BONAFIDE>}\}$,
and $t_{\leq T}$ denotes the concatenated visual and textual context.

Model optimization is performed using the AdamW optimizer. Hyperparameters are selected through Optuna-based optimization with validation EER as the primary objective. For the SmolVLM2-500M-Video-Instruct variant, the learning rate is sampled from $[1\times10^{-6},\,1\times10^{-5}]$, while for the SmolVLM2-2.2B-Instruct variant it is sampled from $[1\times10^{-5},\,1\times10^{-4}]$, based on a set of preliminary experiments. Across both variants, the warmup ratio is sampled from $[0.03,\,0.10]$ of the total training steps, the weight decay coefficient from $[1\times10^{-6},\,1\times10^{-4}]$, the LoRA rank from $\{8,16,32\}$, the scaling factor $\alpha$ from $\{16,32,64\}$, and the LoRA dropout rate from $[0.0,\,0.2]$.

For each hyperparameter configuration, three independent training trials are conducted to reduce sensitivity to random initialization. Each trial is trained for up to $30$ epochs, with evaluation performed every $10$ epochs to monitor both training loss and validation EER. Model selection is based on validation performance; the lowest validation EER during training is retained. After training, only fully merged models are retained, rather than preserving standalone LoRA adapters.

Following Optuna-based hyperparameter optimization, the model checkpoint corresponding to the lowest EER observed on the validation set is selected for subsequent evaluation. For each test sample, posterior probabilities for the \textit{Attack} and \textit{Bona fide} classes are computed from the normalized likelihoods of the corresponding label tokens conditioned on the image–prompt input.

The generative structure shows a strong split between genuine ID-card datasets and synthetic datasets. On genuine data, SmolVLM2 Generative structure produces low EER and maintains low BPCER at strict operating points, whereas on synthetic data, its performance varies sharply by country and can approach near-complete bona fide rejection at strict thresholds.

For the Chile dataset, the generative structure is highly competitive but does not surpass the strongest unimodal model. SmolVLM2-2.2B-Gen reaches 0.93\% EER, which is substantially better than the DenseNet baseline (2.21\% EER), but it is slightly worse than the unimodal SigLIP ablation (0.85\% EER). The same ranking is reflected at strict operating points.

For the Mexico dataset, the benefit of the generative multimodal approach is more visible. SmolVLM2-2.2B-Gen achieves 5.99\% EER, improving over DenseNet (36.77\% EER) and also over the unimodal SigLIP model (9.17\% EER). At stricter operating points, SmolVLM2-2.2B-Gen also maintains low BPCER, whereas SigLIP’s BPCER increases more noticeably. 

Towards model-scale comparison (500M vs 2.2B), in Chile, dataset both generative variants are already low-error (500M 1.96 EER, 2.2B 0.93\% EER). In the investigation of Mexico dataset, however, scaling is decisive: 500M-Gen remains high (25.88\% EER), while 2.2B-Gen drops to 5.99\% EER.

On the synthetic datasets, the generative structure does not provide consistent gains over the baseline or the unimodal ablation, and in the analysis of the Spain dataset behaves as a clear outlier. For example, both the unimodal model and the generative models show extreme failure under strict operating points, and SmolVLM2-Gen also yields BPCER near 100. Compared with DenseNet, the generative model is not uniformly better on synthetic sets either (e.g., Poland: DenseNet 23.58\% EER vs 2.2B-Gen 24.82\% EER).

\paragraph{Discriminative Structure}
In contrast to the generative structure, the discriminative structure replaces the next-token prediction objective with a direct binary classification, while preserving the same frozen vision encoder, prompt design, and LoRA-based parameter-efficient adaptation. Instead of relying on label token likelihoods, the model is trained to produce class-specific scores for Attack and Bona fide directly.

Training is performed using a binary cross-entropy loss over the predicted class probabilities,
\[
\mathcal{L}
= -\big[y \log p_{\text{Attack}} + (1-y)\log p_{\text{Bona\_fide}}\big],
\]
where $y \in \{0,1\}$ denotes the ground-truth label. The loss is computed from a lightweight classification head operating on the final hidden representation at the generation start position, while LoRA adapters remain the only trainable components within the backbone, as in the generative structure.

Hyperparameters are optimized using Optuna, with the optimization objective defined as the validation EER. Unlike the generative structure, optimization in the discriminative setting is driven by class-level logits rather than next-token likelihoods. To reflect the different optimization dynamics of the trainable components, Optuna jointly searches over two independent learning rates: one for the LoRA adapters and one for the classification head. For SmolVLM2-500M-Video-Instruct, the search ranges are $\eta_{\text{lora}} \in [10^{-6}, 10^{-5}]$ and $\eta_{\text{head}} \in [10^{-5}, 10^{-4}]$. For the larger SmolVLM2-2.2B-Instruct, both ranges are shifted upward to $\eta_{\text{lora}} \in [10^{-5}, 10^{-4}]$ and $\eta_{\text{head}} \in [3\times10^{-5}, 3\times10^{-4}]$. All remaining hyperparameters follow the same search space as in the generative structure 
setting.
All hyperparameter optimization, model selection, and evaluation are performed consistently under the generative structure using validation EER as the sole selection criterion.

The performance of the multimodal discriminative structure is reported in
Tables~\ref{tab:pad_Chile_Mexico} and~\ref{tab:pad_Poland_Portugal_Spain}. On genuine datasets in Chile and Mexico, the discriminative structure does not match the strongest baselines. In Chile, both discriminative variants exhibit substantially higher EER and BPCER than DenseNet and SigLIP-SO400M. In Mexico, increasing model capacity from 500M to 2.2B improves discriminative performance, allowing the larger variant to outperform DenseNet; however, it remains clearly inferior to the unimodal model and the generative structure.

On the synthetic datasets, the discriminative structure shows more mixed behavior. On the Poland, its performance is broadly comparable to DenseNet and SigLIP-SO400M, without a consistent advantage. For the Portugal dataset, the discriminative structure, particularly the 500M variant, achieves noticeably lower error rates than both the deep learning baseline and the unimodal ablation. In Spain, however, the discriminative structure fails to generalize effectively, with high EER and rapidly increasing BPCER.
Scaling the discriminative model from 500M to 2.2B generally improves performance on genuine datasets but yields inconsistent gains on synthetic data.

Overall, the multimodal discriminative structure does not outperform the deep learning baseline or the unimodal model on genuine ID Card datasets, and only shows advantages on specific synthetic datasets.

\section{Analysis}
\label{sec:Ana}

The experimental results demonstrate that zero-shot multimodal models are fundamentally incapable of performing the PAD on IDs task without task-specific supervision. Despite extensive pretraining, both the 500M and 2.2B variants operating in a zero-shot setting yield near random performance, with EERs of 50.46\% and 45.26\% on the Chile dataset, respectively. This suggests that general multimodal pretraining models are not enough to capture PAD for ID tasks on its own. Once supervised fine-tuning is applied, however, performance improves dramatically. On the Chile dataset, the generative structure reduces EER from near-random zero-shot performance to 0.93\%, demonstrating that supervised adaptation enables the model to associate visual evidence with explicit authenticity labels.

Across all evaluated models, a clear distinction emerges between performance on genuine ID datasets and synthetic ID datasets. On genuine datasets, both unimodal and multimodal approaches outperform traditional deep learning baselines, whereas results on synthetic datasets (from Poland, Portugal, and Spain) are highly unstable across all modeling strategies. Specifically, this manifests as in contrast to the relatively consistent behavior observed on genuine data, synthetic datasets produce large performance fluctuations across countries and models, including extreme failure cases such as BPCER values reaching 100\% for the Spain dataset. These results indicate that strong performance on genuine ID Cards does not translate reliably to synthetic document domains, regardless of the modeling strategy employed.

When focusing specifically on genuine ID datasets, consistent trends emerge across both controlled and cross-country evaluation settings. For the Chile dataset, where training and testing conditions are aligned, the unimodal model achieves the lowest error rate with an EER of 0.85\%, closely followed by the generative structure at 0.93\%, while the discriminative structure performs substantially worse at 17.08\% EER. Traditional deep learning baselines remain competitive under these compared conditions, achieving EER values below 3\%. This ranking changes significantly for the Mexico dataset, introducing a genuine cross-country domain shift. Under this setting, the generative structure maintains relatively stable performance, achieving an EER of 5.99\%. The unimodal model degrades under cross-country variation, with EER increasing to 9.17\%, while the discriminative structure exhibits a more pronounced performance drop to 19.29\%. Traditional deep learning baselines experience the most severe degradation, with EER rising to 36.77\%, indicating limited robustness to cross-country variation. Taken together, results from the Chile and Mexico datasets show that while multiple approaches perform well under matched conditions, the generative structure demonstrates superior robustness when evaluated on genuine ID Cards from unseen countries, which indicates that the participation of the textual modality increases the generalization ability of the model in PAD on the ID task.

Performance on synthetic ID datasets exposes a critical limitation shared by all evaluated approaches. Across the datasets from Poland, Portugal, and Spain, error rates fluctuate substantially and do not follow the trends observed on genuine data. For the Spain, both unimodal baselines and multimodal models reach extreme failure cases, with BPCER values of 100\%, indicating a systematic mismatch between learned decision cues and synthetic document characteristics. Although isolated instances of good performance exist---such as the discriminative structure achieving an EER of 9.66\% with the Portugal dataset---these results are not consistent across other synthetic datasets. Overall, the instability observed on synthetic ID Cards suggests that synthetic document characteristics differ fundamentally from those present in genuine IDs, limiting the transferability of PAD cues learned from real data.

Finally, scaling effects play a decisive role in managing PAD complexity on genuine ID datasets. On the Mexico dataset, the generative structure with lower model capacity exhibits a high EER of 25.88\%, whereas scaling to the larger variant reduces the error to 5.99\%. This substantial gap indicates that increased capacity is essential for capturing distinctions between bona fide IDs and presentation attacks under cross-country variation. However, increased capacity does not mitigate instability on synthetic datasets, reinforcing the conclusion that synthetic ID Cards remain a fundamentally unresolved challenge for all evaluated modeling strategies.

\section{Conclusion}
\label{sec:conclusion}

%%The experimental results demonstrate that, although compact multimodal models exhibit strong generalization capacity after adaptation, they are fundamentally incapable of performing PAD on ID Cards in a zero-shot setting. Once supervised fine-tuning is applied, however, effective PAD behavior emerges. In particular, the generative structure leads to substantially lower error rates on genuine ID datasets after supervised fine-tuning.

%%The experimental results highlight the importance of model size for PAD on genuine ID datasets. Scaling the generative structure from lower to higher capacity variants substantially reduces error under cross-country evaluation, indicating that model capacity directly affects generalization to unseen genuine datasets. In contrast, increased model capacity does not provide consistent benefits on synthetic ID datasets. Across Poland, Portugal, and Spain, all evaluated approaches exhibit highly unstable behavior, with performance varying sharply across countries and, in some cases, collapsing entirely. Larger models do not systematically reduce error on synthetic data, frequently failing to prevent extreme outcomes such as BPCER values reaching 100\%.

A central finding of this study is that, on genuine ID datasets, the generative structure demonstrates superior robustness to cross-country domain shifts compared to discriminative structures, unimodal models, and traditional deep learning baselines. While most approaches achieve strong performance under matched training and testing conditions on genuine data, the generative structure consistently mitigates the performance degradation observed when evaluating genuine ID Cards from unseen countries. In contrast, discriminative structures, unimodal models, and conventional baselines exhibit significantly larger error increases under cross-country variation on genuine datasets. This indicates that incorporating textual modality allows the multimodal model to generalize more effectively across genuine ID datasets from unseen countries.

Despite these improvements on genuine ID Card datasets, the results expose a critical limitation in current PAD evaluation practices involving synthetic data. All evaluated modeling strategies exhibit unstable and often catastrophic performance on synthetic ID datasets, indicating a fundamental mismatch between learned PAD cues from real documents and the visual characteristics of synthetic IDs. These findings suggest that, while generative multimodal structures provide a reliable solution for PAD in realistic, real-world scenarios, synthetic datasets remain an unresolved challenge. 
\vspace{-0.3cm}

\section*{Acknowledgements}
This work was partially supported by EU Horizon under G.A. EINSTEIN (101121280) and CarMen (101168325), and the German Federal Ministry of Education and Research and the Hessian Ministry of Higher Education, Research, Science and the Arts within their joint support of the National Research Center for Applied Cybersecurity ATHENE.

%%%%%%%%%%%%%%%%%%%%%%%%%%%%%%
% BIBLIOGRAPHY
%%%%%%%%%%%%%%%%%%%%%%%%%%%%%%
\bibliographystyle{IEEEtran}
\bibliography{main.bib}
\vspace{-0.3cm}
%
%%%%%%%%%%%%%%%%%%%%%%%%%%%%%%
% BIOGRAPHIES
%%%%%%%%%%%%%%%%%%%%%%%%%%%%%%
\begin{IEEEbiography}[{\includegraphics[width=1in,height=1.25in,clip,keepaspectratio]{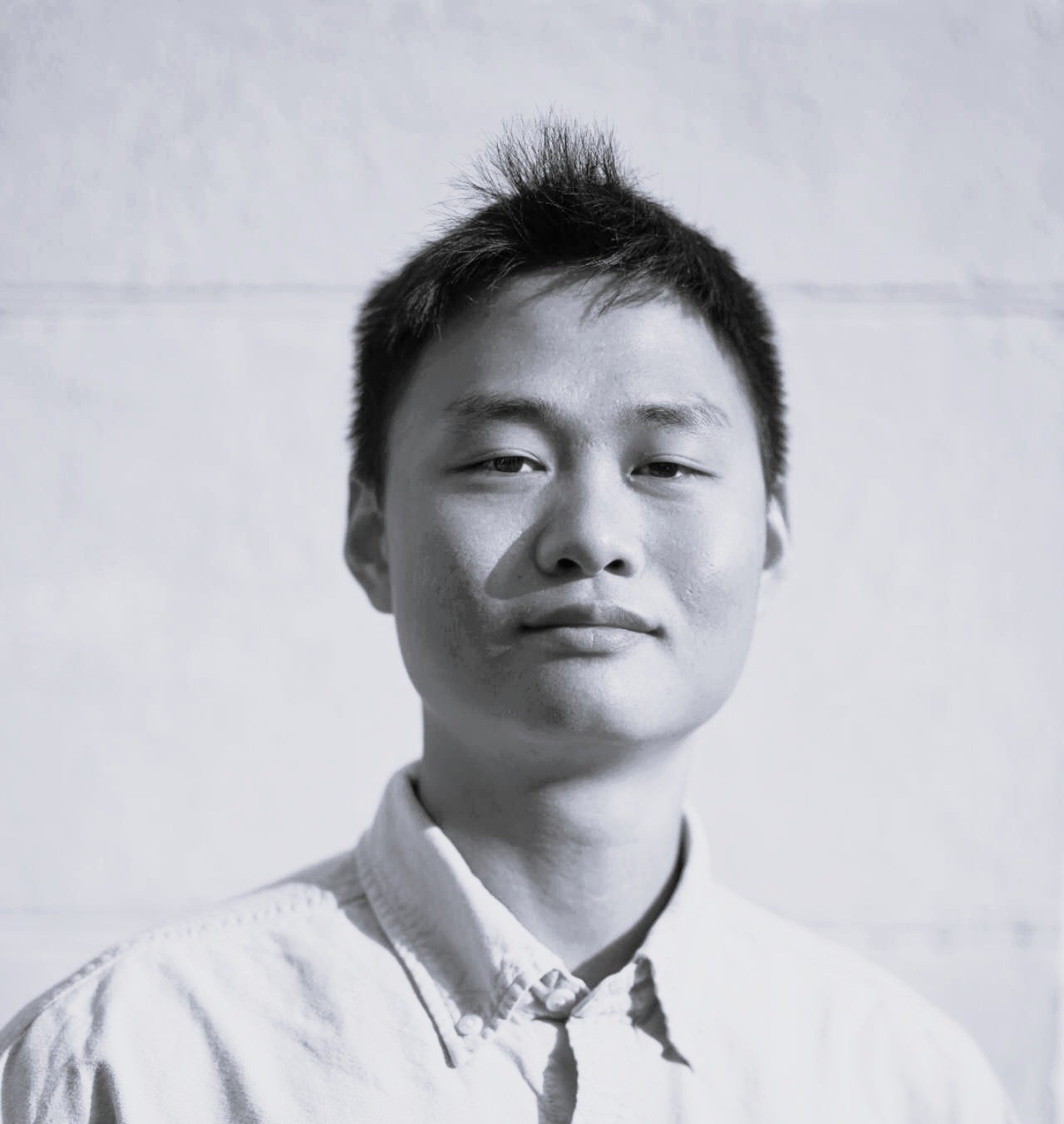}}]{Qingweng Zeng} received a B.Sc. degree in Data Science from the Institute of Disaster Prevention, China, in 2023. He is pursuing his M.Sc. degree in Human-Centred Artificial Intelligence at the Technical University of Denmark. His research focuses on leveraging state-of-the-art AI technologies to advance the field of biometrics.
\end{IEEEbiography}
\vspace{-0.5cm}

\begin{IEEEbiography}[{\includegraphics[width=0.9in,height=1.25in,clip,keepaspectratio]{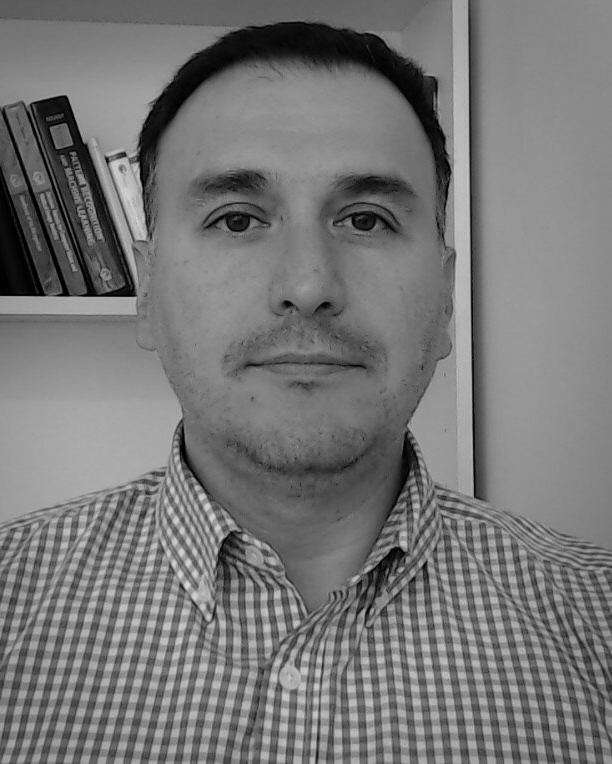}}]{Juan Tapia} received the P.E. degree in electronics engineering from Universidad Mayor, in 2004, and the M.S. and Ph.D. degrees in electrical engineering from the Department of Electrical Engineering, Universidad de Chile, in 2012 and 2016, respectively. In addition, he spent one year of internship with the University of Notre Dame. In 2016, he received the Award for Best Ph.D. Thesis. From 2016 to 2017, he was an Assistant Professor at Universidad Andrés Bello. From 2018 to 2020, he was the Research and Development Director for the electricity and electronics area with INACAP, Universidad Tecnológica de Chile, the Research and Development Director of TOC Biometrics Company, and an International Advisor on biometrics for face, iris applications and forensic/tampering ID-card detection. He is currently an Entrepreneur and a Senior Researcher with Hochschule Darmstadt (H-DA), leading EU projects, such as iMARS, EINSTEIN and CarMen. His main research interests include pattern recognition and deep learning applied to iris biometrics, morphing, feature fusion, and feature selection. He serves as a reviewer for several journals and conferences. He is on behalf of the German DIN as a Member of the ISO/IEC Sub-Committee 37 on biometrics.
\end{IEEEbiography}
\vspace{-0.5cm}

\begin{IEEEbiography}[{\includegraphics[width=1in,height=1.25in, clip,keepaspectratio]{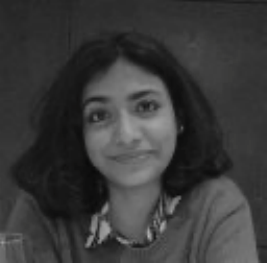}}]{Sneha Das} {\,} (Member, IEEE)  is a tenure-track Assistant Professor in Statistics and Data Analysis at the Department of Applied Mathematics and Computer Science, Technical University of Denmark (DTU), Copenhagen, Denmark. She received her Ph.D. in 2021 from Aalto University, Finland, within the Speech and Language Technology group. Her current research spans AI-safety, low-resource machine learning, and multi-sensory signal processing (including speech and audio, biosignals and visual modalities) with applications in healthcare and education. She works at the intersection of signal processing, machine learning and statistics.  
\end{IEEEbiography}
\vspace{-0.5cm}

\begin{IEEEbiography}[{\includegraphics[width=0.9in,height=1.25in,clip,keepaspectratio]{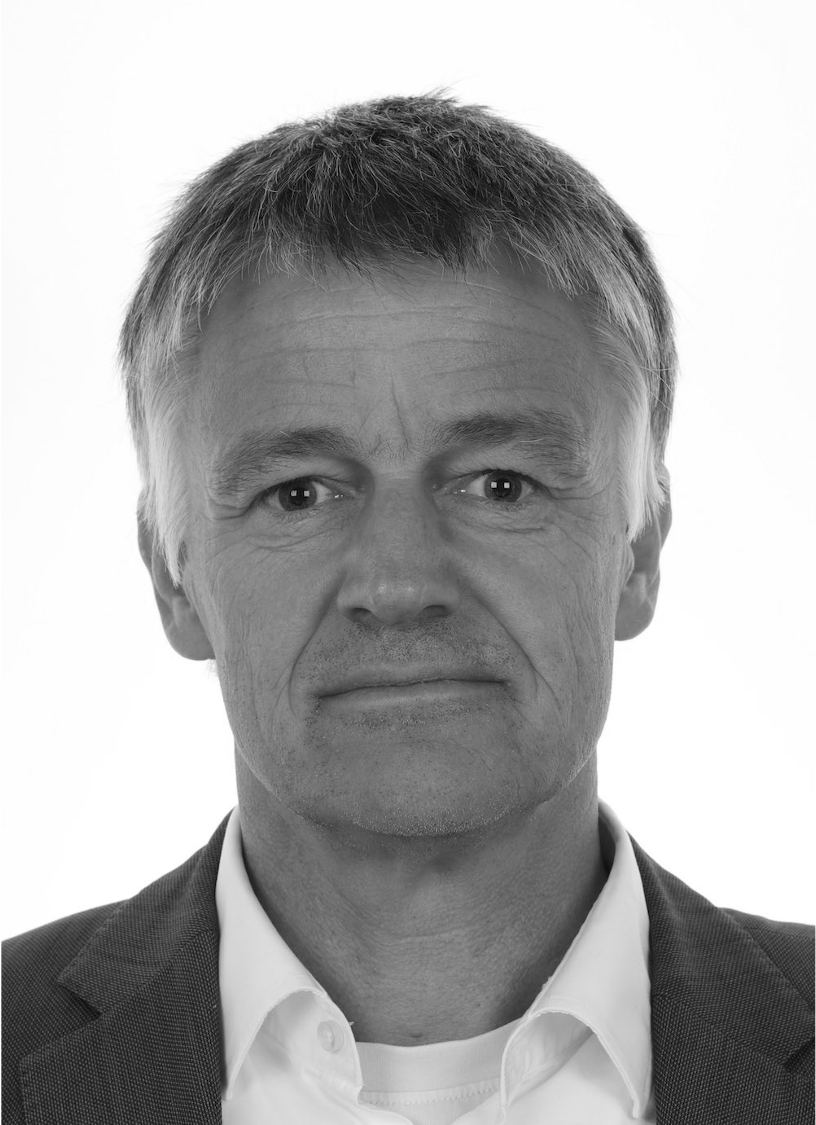}}]{Christoph Busch} is a member of the Department of Information Security and Communication Technology (IIK) at the Norwegian University of Science and Technology (NTNU), Norway. He holds a joint appointment with the computer science faculty at Hochschule Darmstadt (HDA), Germany. Further, he lectures the course Biometric Systems at Denmark’s DTU since 2007. On behalf of the German BSI he has been the coordinator for the project series BioIS, BioFace, BioFinger, BioKeyS Pilot-DB, KBEinweg and NFIQ2.0. In the European research program, he was the initiator of the Integrated Project 3D-Face, FIDELITY and iMARS. Further, he was/is partner in the projects TURBINE, BEST Network, ORIGINS, INGRESS, PIDaaS, SOTAMD, RESPECT and TReSPAsS. He is also principal investigator at the German National Research Center for Applied Cybersecurity (ATHENE). Moreover, Christoph Busch is co-founder and member of the board of the European Association for Biometrics (www.eab.org) which was established in 2011 and assembles in the meantime more than 200 institutional members. Christoph co-authored more than 700 technical papers and has been a speaker at international conferences. He is a member of the editorial board of the IET journal.
\end{IEEEbiography}

\vfill

\end{document}